\documentclass[letterpaper, 10 pt, journal, twoside]{IEEEtran}

\IEEEoverridecommandlockouts                              

\usepackage{graphicx}
\usepackage{hyperref}
\usepackage{tikz}
\usepackage{makecell}
\usepackage{diagbox}
\usepackage{orcidlink}
\usepackage{mathtools}
\usepackage{svg}
\usepackage[ruled,vlined]{algorithm2e}
\LinesNumbered
\SetArgSty{textup}
\hypersetup{
    colorlinks=true,
}
\usepackage{amsfonts}
\usepackage{amsmath}
\usepackage[thinlines]{easytable}
\usepackage{hhline}
\usepackage{stackengine}
\newcommand\xrowht[2][0]{\addstackgap[.5\dimexpr#2\relax]{\vphantom{#1}}}
\usepackage{cite}
\usepackage{booktabs}
\usepackage[font=footnotesize,labelsep=period]{caption}[2022/02/20]
\usepackage{subcaption}
\usepackage{caption}
\usepackage{float}
\usepackage{amsmath}
\usepackage{amssymb}
\usepackage{balance}

\usepackage{bm}

\newcommand{\xxnote}[3]{}
\ifx\hidenotes\undefined
  \usepackage{color}
  \renewcommand{\xxnote}[3]{\color{#2}{#1: #3}}
\fi

\definecolor{review}{cmyk}{0,0,0,1}
\definecolor{reviewbs}{rgb}{0.5,0.3,0.3}
\definecolor{done}{cmyk}{0,0,0,1}

\DeclareMathOperator*{\argmin}{arg\,min}

\newcommand*{\vecbf}[1]{\bm{#1}} 
 
\newcommand*{\pxminus}{x_k} 
\newcommand*{\pxplus}{x_{k+1}} 
\newcommand*{\pyminus}{y_k} 
\newcommand*{\pyplus}{y_{k+1}}

\begin{document}
\title{
Hierarchical Planning for Long-Horizon Multi-Target Tracking Under\\Target Motion Uncertainty
}

\author{Junbin Yuan$^{1}$, Brady Moon$^{2}$, Muqing Cao$^{2}$, and Sebastian Scherer$^{2}$ 
\thanks{This work is supported by the Office of Naval Research under Grant No. N00014-21-1-2110 and Contract No. N6833522C0179.. Additionally, this work was supported by the National Science Foundation Graduate Research Fellowship under Grant No. DGE1745016.}
\thanks{$^{1}$Author is with the Mechanical Engineering Department at Carnegie Mellon University, Pittsburgh, PA, USA {\tt\footnotesize  junbiny@andrew.cmu.edu}}
\thanks{$^{2}$Authors are with the Robotics Institute, School of Computer Science at Carnegie Mellon University, Pittsburgh, PA, USA{\tt\footnotesize  \{bradym, muqingc, basti\}@andrew.cmu.edu}}
}


\maketitle

\begin{abstract}
Achieving persistent tracking of multiple dynamic targets over a large spatial area poses significant challenges for a single-robot system with constrained sensing capabilities.
As the robot moves to track different targets, the ones outside the field of view accumulate uncertainty, making them progressively harder to track. An effective path planning algorithm must manage uncertainty over a long horizon and account for the risk of permanently losing track of targets that remain unseen for too long.
However, most existing approaches rely on short planning horizons and assume small, bounded environments, resulting in poor tracking performance and target loss in large-scale scenarios.
In this paper, we present a hierarchical planner for tracking multiple moving targets with an aerial vehicle. To address the challenge of tracking non-static targets, our method incorporates motion models and uncertainty propagation during path execution, allowing for more informed decision-making. We decompose the multi-target tracking task into sub-tasks of single target search and detection, 
and our proposed pipeline consists a novel low-level coverage planner that enables searching for a target in an evolving belief area, and an estimation method to assess the likelihood of success for each sub-task, making it possible to convert the active target tracking task to a Markov decision process (MDP) that we solve with a tree-based algorithm to determine the sequence of sub-tasks.
We validate our approach in simulation, demonstrating its effectiveness compared to existing planners for active target tracking tasks, and our proposed planner outperforms existing approaches, achieving a reduction of 11--70\% in final uncertainty across different environments. We also conduct a real-robot experiment to verify our method's practical deployability. The project website is at \href{https://yuanjunbin.github.io/hptracking}{https://yuanjunbin.github.io/hptracking/}.
\end{abstract}

\begin{IEEEkeywords}
Planning under Uncertainty; Task and Motion Planning; Motion and Path Planning
\end{IEEEkeywords}

\IEEEpeerreviewmaketitle

\section{Introduction}

\IEEEPARstart{T}{arget} dtracking has a growing number of real-world applications, including aerial surveillance \cite{airsurveillance_1}, vehicle tracking on road networks \cite{roadnet}, and wildfire monitoring \cite{wildfire_example}. In many of these scenarios, robots must track multiple targets dispersed across large, unbounded environments, moving with uncertainty over extended time horizons. These challenges demand intelligent path planning strategies to ensure persistent and efficient target coverage.

The long-horizon multi-target tracking problem presents multiple challenges. First, moving targets create a dynamic environment where most data gathering path planning algorithms, designed for static settings, are no longer effective. For example, existing planning techniques like graph search would have to handle a graph where all edges and nodes change over time. 
Second, when the number of targets exceeds the number of robots, continuous tracking of all targets may not be possible \cite{multi_agent_vis}. This necessitates a planner capable of strategically switching between targets to maximize overall tracking success.
Therefore, the problem presents challenges in both high-level decision-making (which targets to track and when) and low-level path planning (how to track and reacquire a target).
However, most existing approaches rely on short planning horizons and are verified only on small workspaces, failing to account for the long-term impact of decisions,
leading to poor tracking performance and lost targets.



\begin{figure}[t]
    \centering
    \includegraphics[trim={0.431cm 1.4224cm 0.4318cm 1.4224cm},clip,width=\columnwidth]{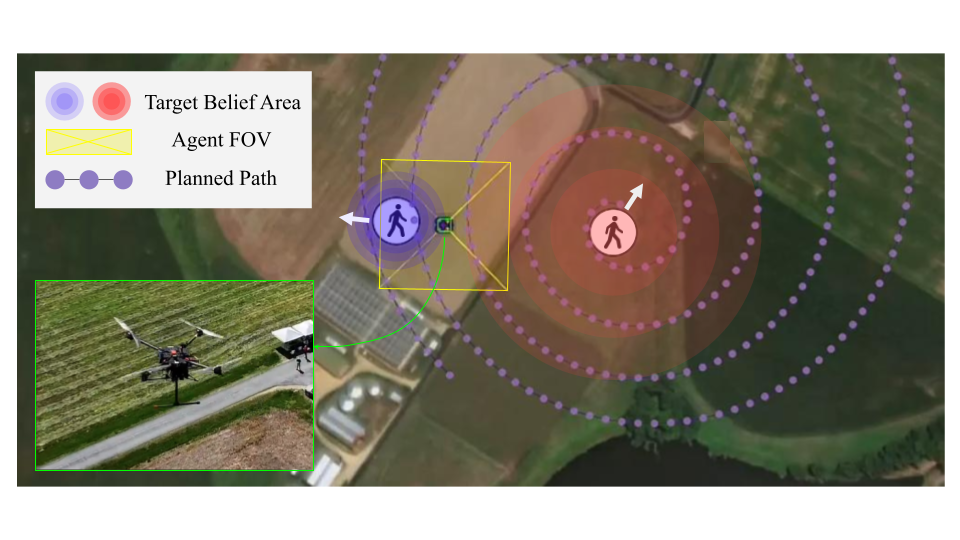} 
    \caption{A UAV tracking two moving targets in an open environment. Each target's belief region expands over time due to motion uncertainty and contracts upon detection. In this snapshot, the UAV has just detected the purple target and is initiating a coverage path to search for the red target.}
    \label{fig: full_overview}
    \vspace{-0.3cm}
\end{figure}

\begin{figure*}[t]
    \centering
    \includegraphics[trim={0.25cm 5cm 5cm 1.75cm},clip,width=\textwidth]{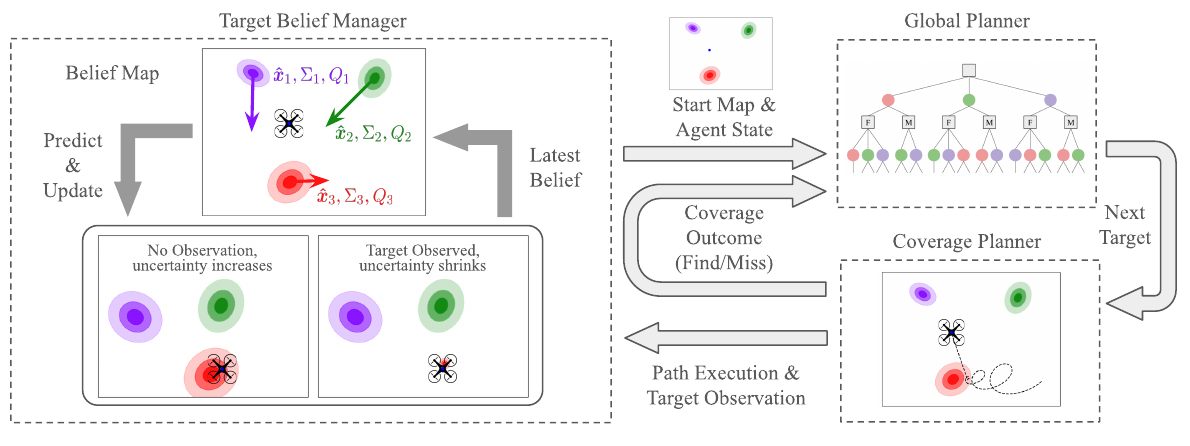}
    \caption{An overview of the system. The target belief manager updates targets’ beliefs using prediction models \eqref{eq:prediction_model}, and applies observation, shrinking the uncertainty whenever a target is observed. It provides information for the global planner to decide the next target to pursue, then a coverage planner generates a path accordingly. }
    \label{fig: overview_fig}
    \vspace{-0.3cm}
\end{figure*}

In this paper, we focus on using a single unmanned aerial vehicle (UAV) to track multiple targets spread in a large and open area with uncertain kinematic models.
We present a hierarchical planning solution consisting of two main components: a high-level target priority planner and a low-level coverage path planner.
The high-level planner utilizes a Monte Carlo tree search to find the order of targets to visit, which results in the maximum expected reduction in uncertainty of target locations when the time budget depletes.  
The low-level planner computes a spiral path that efficiently covers the growing ellipse region representing the probable location of a target.
The deterministic path allows us to estimate the probability and time taken to find a target, which is used in the Monte Carlo tree search.
Our approach is tested in extensive numerical simulations and shows significant improvement over existing methods. The main contributions are:
\begin{itemize}
    \item A low-level shifting spiral coverage path planner that searches for a target in a dynamically propagating belief region.
    \item A coverage search success estimator that predicts the probability of finding the target and estimated search time, enabling the formulation of multi-target tracking as a Markov decision process planning problem.
    \item A high-level task allocation planner using Monte-Carlo tree search (MCTS) to determine target visitation sequence, minimizing overall uncertainty across all targets upon budget depletion.
    \item Comprehensive simulation shows that our approach outperforms existing approaches, achieving a reduction of 11--70\% in final uncertainty across different environments.
    \item Real-world flight experiments verify the practicality of the approach. 
\end{itemize}


\section{Related Work}\label{sec:related}

Target tracking using mobile robots has been actively investigated by the research community.
Particularly, many related works have been proposed within the broader domain of target tracking \cite{khan2016cooperative,robin2016multi}, but they differ substantially in their task settings and problem formulations.
In \cite{Spletzer2003DynamicSP}, the target tracking problem is defined as optimizing the estimation of target position with combined measurements from multiple robots.
Strategies for cooperative discovering and tracking of multiple moving targets are studied in \cite{1997CMOMMT} and its follow-up works \cite{parker2002distributed,kolling2007cooperative}, where the task is to maintain persistent visibility of moving target using robots with sensors.

With regard to cases where targets outnumber robots, 
\cite{banfi2015fair} maintains target belief using a Bayesian filter and formulates a multi-objective integer linear program to achieve even coverage of targets.
To maintain computational feasibility, optimization is solved with a short planning horizon and discrete small workspace.
The trade-off between the quality of tracking each target and the number of targets tracked is studied in \cite{multi_agent_vis}.
Some studies apply the Probability Hypothesis Density (PHD) filter to update and predict the belief over target states, and propose various search-and-track strategies using greedy heuristic  \cite{philip2017detecting,Papaioannou2021}, 
  Voronoi-cell-based control \cite{dames2020distributed}, 
 Thompson sampling \cite{Arundhati2024decentralized}, and Particle Swarm Optimization \cite{xin2022comparing}. 
A recent work \cite{van2024multi} formulates a multi-objective optimization problem including both search and tracking-related objectives, and solves it using a greedy solution. A following work created a unified objective for search and track \cite{santos2025unisat}, eliminating the need for a discrete, gridded search space.

Several works have studied the specific problem of tracking multiple targets using a single robot, which is closer to our task. The task of active target tracking \cite{5735231} is formulated with state propagation and observation model. \cite{roadnet} considers the targets' state propagation constrained to the roadways and provides a particle filter to update the belief of targets and a receding horizon planning method with a look-ahead step where particles are propagated forward. 
For targets moving in open space, \cite{8260881} developed Anytime Reduced Value Iteration (ARVI) for active information gathering, and \cite{8968147} proposed a task-specific heuristic for A* planning to address the active target tracking problem. 
Some learning-based methods have also been proposed for active target tracking, including \cite{8968173} and \cite{jeong2021deep} that are based on Assumed Density Filtering Q-learning (ADFQ) \cite{10.5555/3367243.3367402}.
However, there are drawbacks that multiple trainings have to be done separately for different numbers of targets, and only the results of tracking one and two targets are presented in \cite{jeong2021deep}. 
A more recent work developed a policy network with attention layer \cite{pmlr-v211-yang23a} that handles a range of targets to track.

These works address an active target tracking task closely related to ours, but their approaches share a common limitation of short horizons and small environments, causing them to struggle in large scenarios with sparse targets that produce intermittent observation. To address this difficulty, we propose a hierarchical planning pipeline that decomposes the active target tracking task into a sequence of single-target detection sub-tasks, improving both concentration and planning efficiency.

\section{Problem Definition}\label{sec:problem}
We consider a $2$D open space in which a single robot tracks $n$ targets with the knowledge of targets' initial estimates, uncertainties, and kinematics model. 
Each target $i$'s state at time $k$ is denoted as $\vecbf{x}_{i,k} = [x_{i,k}, y_{i,k}, v^x_{i,k}, v^y_{i,k}]^{\top}$ consisting of its position and velocity.
The initial state estimates of the targets are given as $\hat{\vecbf{x}}_{i,0}$, and the initial uncertainty is described as a covariance matrix $\Sigma_{i,0}$, $i\in[1,n]$. 
The targets follow a simple motion model with process noise: 
\begin{align*}
    \vecbf{x}_{i,k} = F \vecbf{x}_{i,k-1} + W_i, \quad i\in[1,n],
\end{align*}
where the state transition matrix $F$ follows a constant velocity model, $W_i$ is a zero-mean random process noise, $W_i=[w^x, w^y, w^{vx}, w^{vy}]^\top\sim\mathcal{N}(\vecbf{0}, Q_i)$, $\tau$ is the discrete time step.

When the robot observes a target within it circular sensor footprint with a diameter $w$, the uncertainty of the target reduces according to the observation covariance, otherwise, the uncertainty keeps increasing over time with
\begin{equation}
\begin{split}
    \hat{\vecbf{x}}_{i,k} &= F \hat{\vecbf{x}}_{i,k-1}, \\
    \Sigma_{i,k} &= F \Sigma_{i,k-1} F^{\top} + Q_i.
\label{eq:prediction_model}
\end{split}
\end{equation}

We define our multi-target tracking problem as finding a robot trajectory $\mathcal{T}$ that minimizes the uncertainty metric $U(\mathcal{T})$ while subject to the budget equality constraint $B$
\begin{equation*}
    \mathcal{T}^* = \argmin_{\mathcal{T}  \in \mathbb{T}} U(\mathcal{T})
    \text{ s.t.\ $C(\mathcal{T}) = B$},
\end{equation*}
where $\mathbb{T}$ denotes the set of all possible trajectories, and $C(\mathcal{T})$ denotes the cost of executing trajectory $\mathbb{T}$, which is constrained by the total available budget $B$. 
In this work, $C(\mathcal{T})$ is the total duration of the trajectory.

The overall track uncertainty $U$ can be defined in various ways, considering average or final state uncertainty as well as uncertainty of specific state variables. In this work, we define $U$ as the covariance of the position estimates at the end of the budget and adopt a similar uncertainty metric as \cite{jeong2021deep}, i.e.,
\begin{equation*}
    U(\mathcal{T}) = \sum_{i \in \text{Targets}} \log (\det \Sigma^{xy}_{i,B}).
\end{equation*}


\section{Methodology}\label{sec:method}
The overview of our approach is shown in Fig. \ref{fig: overview_fig}.
A Target Belief Manager updates targets’ beliefs using prediction models \eqref{eq:prediction_model}, and applies observation, shrinking the uncertainty whenever a target is observed. 
Our hierarchical planning framework comprises two levels of planning: A coverage path planning that searches for a tracked target in an evolving area of uncertainty, and global planning that determines which target the agent should pursue next. 



\begin{figure}[t]
     \centering
     
     \begin{subfigure}[b]{0.55\columnwidth} 
         \centering
         \includegraphics[trim={5cm 0cm 4.6cm 0cm},clip,width=\textwidth]{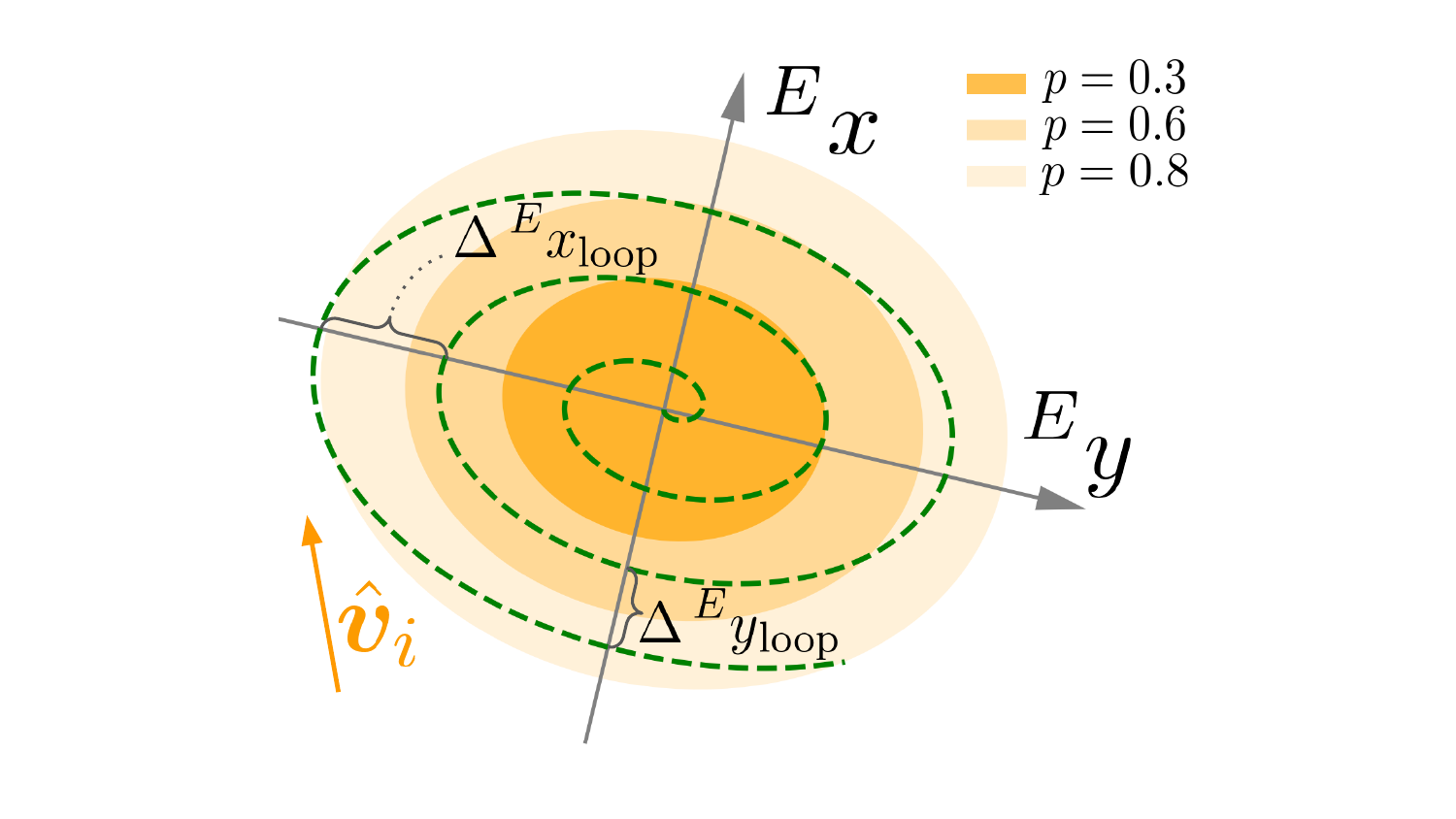}
         \caption{}
         \label{fig: coverage static}
     \end{subfigure}
     \hfill
     \begin{subfigure}[b]{0.43\columnwidth}
         \centering
         \includegraphics[trim={1.5cm 0cm 0cm 0cm},clip,width=\textwidth]{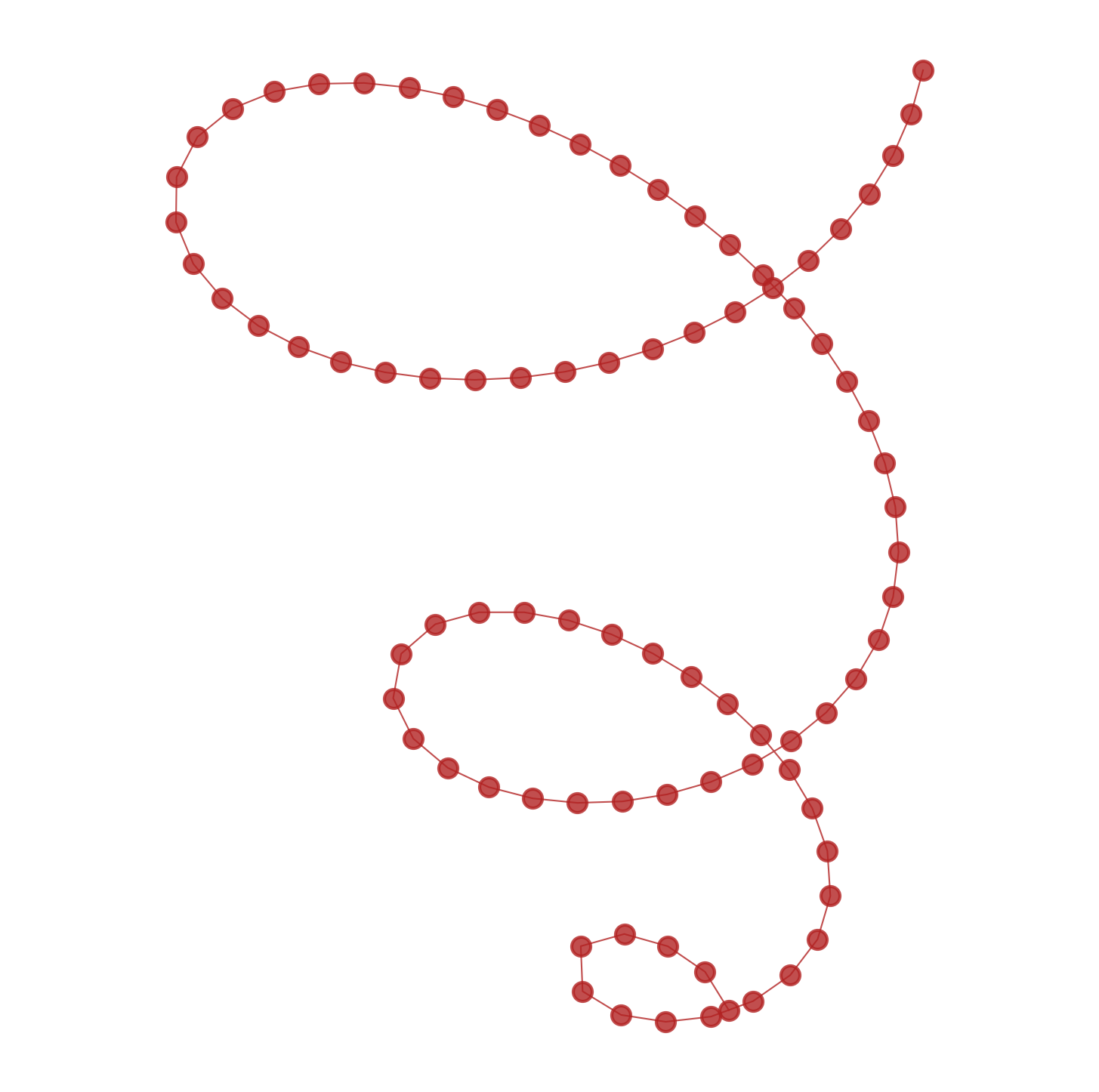}
         \caption{}
         \label{fig: coverage full}
     \end{subfigure}

     \vskip\baselineskip 
     \begin{subfigure}[b]{0.55\columnwidth}
         \centering
         \includegraphics[trim={0cm 0.5cm 0cm 0.5cm},clip,width=\textwidth]{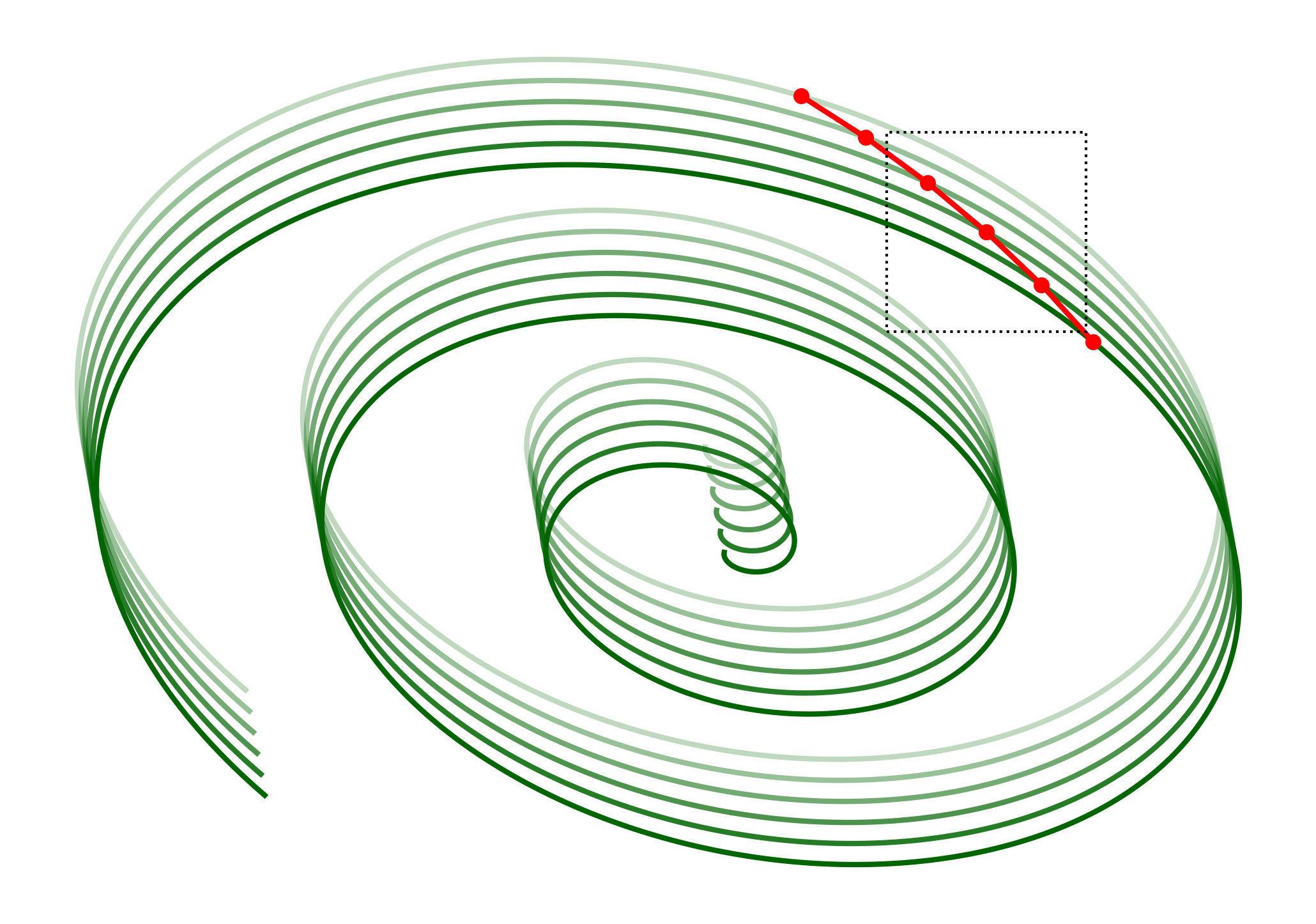}
         \caption{}
         \label{fig: coverage shift}
     \end{subfigure}
     \hfill
     \begin{subfigure}[b]{0.43\columnwidth}
         \centering
         \includegraphics[trim={5.5cm 1cm 5.5cm 1cm},clip,width=\textwidth]{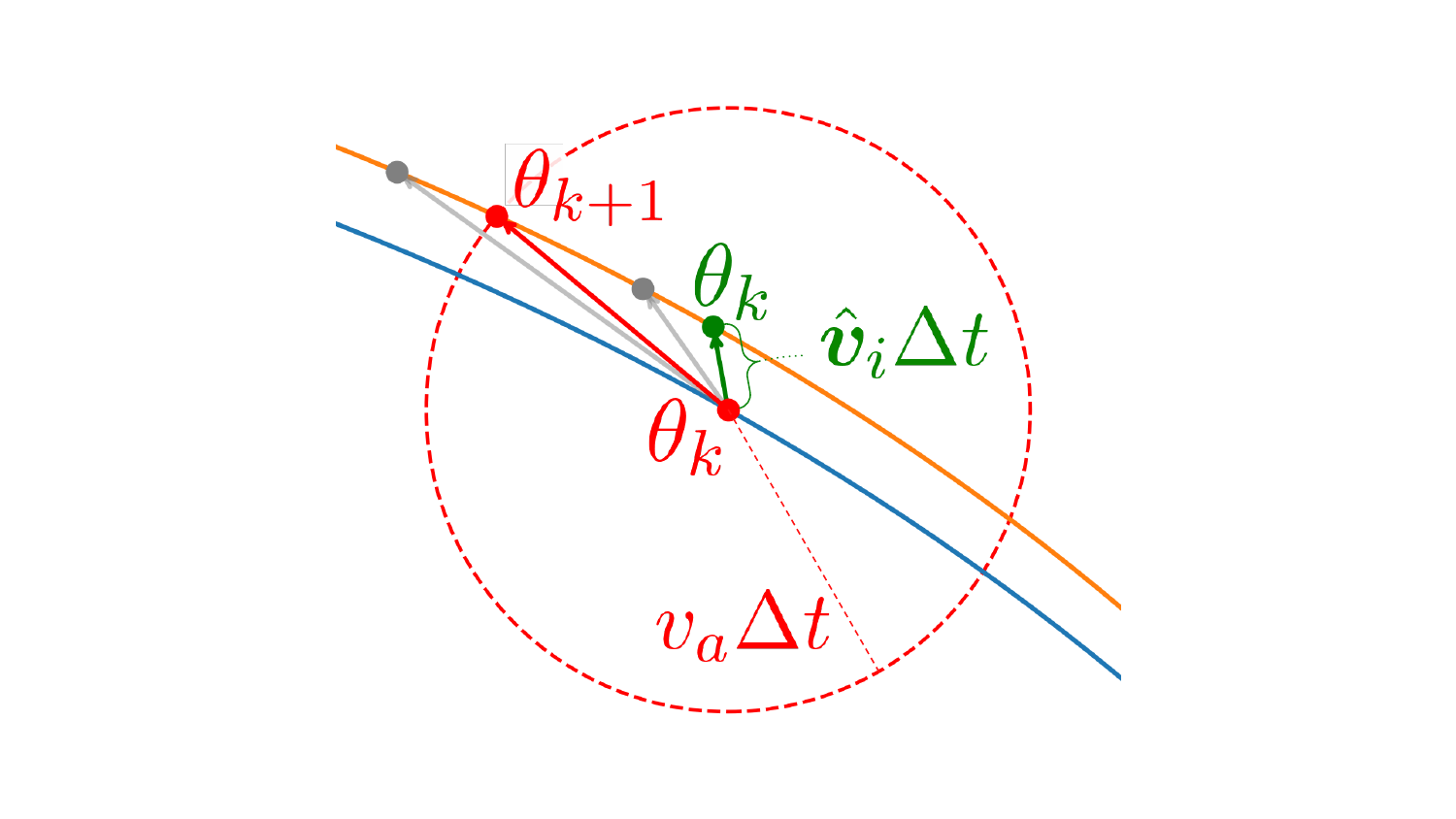}
         \caption{}
         \label{fig: coverage step}
     \end{subfigure}

    \caption{An example coverage path over a moving target. We use the static elliptic spiral (Fig. \ref{fig: coverage static}) to generate a path with equal distance between waypoints (Fig. \ref{fig: coverage full}) to cover a moving target belief area. The path is extended by finding the next waypoint on the shifted static spiral (Fig. \ref{fig: coverage shift}). In each step, a line search from $\theta_k$ acquires $\theta_{k+1}$ that ensures the agent's displacement matches its constant speed (Fig. \ref{fig: coverage step}).}
             
    \label{fig: coverage four}
\end{figure}

\subsection{Coverage Path Planner}
\label{sec: coverage path planner}
The uncertain position of a target is represented as an ellipse of possible locations defined by the top-left $2\times2$ submatrix of the covariance matrix $\Sigma$.  
As the prediction process in \eqref{eq:prediction_model} propagates the target estimate, this uncertainty grows over time, causing the ellipse to expand while following the predicted motion of the target. As illustrated in Fig.~\ref{fig: coverage four}, we implemented a shifting spiral coverage path planner to search for the target in this propagating area of uncertainty.

We define an elliptic spiral path with angular coordinate $\theta_k$ and radial coordinate $r_k$, in the reference frame $\{E\}$ centered at the target $\hat{\vecbf{x}}_{i,k}$ at time $k$ after the coverage starts, and axes directions aligned with the semi-axes of the ellipse: 
\begin{equation}
\begin{split}
    r(\theta&_k) = a + b \theta_k, \\
    \prescript{E}{}{x_k} = r(\theta_k) A \cos&(\theta_k), \quad \prescript{E}{}{y_k} = r(\theta_k) B \sin(\theta_k),
\label{eq:static_spiral}
\end{split}
\end{equation}
where $a,b, A, B$ are the geometric coefficients of the spiral path. 
In each loop with $2\pi$ change in phase $\theta$, the gap on both directions of the semiaxes is (Fig. \ref{fig: coverage static})

\begin{equation*}
    \Delta \prescript{E}{}x_\text{loop} = 2 \pi b A, \quad
    \Delta \prescript{E}{}y_\text{loop} = 2 \pi b B.
\end{equation*}
To densely cover the area, we want to ensure that the gap between the spiral loops is equal to the width of the sensor field of view $w$, i.e., $\Delta \prescript{E}{}x_\text{loop}\leq w,\Delta \prescript{E}{}y_\text{loop}\leq w$. 
We set  
\begin{equation*}
    b = \frac{w}{2 \pi \max(A, B)}
\end{equation*}
to fully utilize the sensor width, and set $a$ with $\theta_0$ such that $r(\theta_0)=0$ to start the coverage from the center of the elliptic area.
We also set $A = \sqrt{\lambda_1}, \quad B = \sqrt{\lambda_2}$, where $\lambda_1, \lambda_2$ are the eigenvalues of the covariance matrix representing the position uncertainty at the beginning of the coverage, $\Sigma^{xy}$. 
The direction of the semi-axes is described by the rotation of the ellipse area, $R$, parameterized by the normalized eigenvectors $\bm{u}_1, \bm{u}_2$ of $\Sigma^{xy}$ corresponding to the two eigenvalues vertical to each other:
\begin{equation*}
    \quad R = 
    \begin{bmatrix}
        u^x_{1} & u^x_{2} \\
        u^y_{1} & u^y_{2}
    \end{bmatrix}.
\end{equation*}
Therefore, the coordinates of the spiral path in the world frame can be expressed as 
\begin{equation}
    \begin{bmatrix}
        x_k \\
        y_k
    \end{bmatrix}
    =
    \begin{bmatrix}
        x_0 \\
        y_0
    \end{bmatrix}
    +
    R
    \begin{bmatrix}
        \prescript{E}{}{x_k} \\
        \prescript{E}{}{y_k}
    \end{bmatrix}
    +
    \begin{bmatrix}
        \hat{v}_x \\
        \hat{v}_y
    \end{bmatrix}
    k\tau
\label{eq: shifting spiral}
\end{equation}

Such a spiral path covers the expanding ellipse area in the reference frame centered at $\hat{x}_{k}$, which starts from $[x_0, y_0]^\top = [\hat{x}_{i, 0}, \hat{y}_{i, 0}]^\top$ when the coverage starts, and the center keeps moving at a {constant speed} $\hat{\vecbf{v}}_{i} = [\hat{v}^x_i, \hat{v}^y_i]^{\top}$. However, computing the spiral path with the linear motion of the target is nontrivial. In the following, we developed a numerical method to generate a viable path.

Assuming the robot follows a shifting spiral path at speed $v_a$, which must be greater than the estimated speed of the target $||\hat{\vecbf{v}}_{i}||$, the fixed spiral path described by \eqref{eq:static_spiral} shifts by a distance of $\hat{\vecbf{v}}_i \tau$ in its reference frame between consecutive time steps $k$ and $k+1$. 
Meanwhile, the robot's angular coordinate $\theta_{k}$ must increase slightly to $\theta_{k+1}$, resulting in a position change within the reference frame $\{E\}$. 
By combining these two motions and using the center of $\{E\}$ at time $k$ as the reference frame position, we obtain
\begin{equation*}
\begin{split}
    \begin{bmatrix}
        x_k \\
        y_k
    \end{bmatrix}
    &=
    R
    \begin{bmatrix}
        r(\theta_k) A \cos (\theta_k) \\
        r(\theta_k) B \sin (\theta_k)
    \end{bmatrix} \\
    \begin{bmatrix}
        x_{k+1} \\
        y_{k+1}
    \end{bmatrix}
    &=
    R
    \begin{bmatrix}
        r(\theta_{k+1}) A \cos (\theta_{k+1}) \\
        r(\theta_{k+1}) B \sin (\theta_{k+1})
    \end{bmatrix}
    +
    \begin{bmatrix}
        \hat{v}_x \\
        \hat{v}_y
    \end{bmatrix}
    \tau
\end{split}
\end{equation*}
With a fixed agent speed $v_a$, the distance between these positions should be equivalent to $v_a \tau$. Therefore, we can form a function
\begin{equation*}
    f(\theta_{k+1}) = (\pxplus - \pxminus)^2 + (\pyplus - \pyminus)^2 - (v_a \tau)^2.
\end{equation*}

To determine $\theta_{k+1}$ for the next time step, we seek the smallest root of $f(\theta_{k+1}) =0$ that is greater than $\theta_k$, given the current phase $\theta_k$, as illustrated in Fig.~\ref{fig: coverage step}. By formulating this as a root-finding problem, we can apply numerical methods such as bisection to iteratively solve for $\theta_{k+1}$. This allows us to extend the shifting spiral path using \eqref{eq: shifting spiral}, ensuring effective coverage of the moving and expanding uncertainty ellipse.

The shifting spiral path can be extended infinitely, but there will be a time $t_\text{cutoff}$ beyond which further spiral coverage doesn't enhance the likelihood of finding the target, which we will discuss in the following subsection.


\subsection{Estimation of Coverage Performance}
\label{sec: coverage estimation}

Before deciding which target to pursue with the coverage planner, we must first be able to estimate key coverage outcomes: (1) the probability of successfully finding the target within $k$ time steps after coverage begins, given by $P_\text{cdf}(k) = P(\text{Find} | t \leq k)$, (2) the cutoff time $t_\text{cutoff}$, beyond which further coverage is considered ineffective, and (3) the overall probability of finding the target with the coverage planner, defined as $p_\text{max} = P_\text{cdf}(t_\text{cutoff})$. To achieve this, we developed an outcome estimator for the coverage planner.



\begin{figure}[t]
    \centering
    \includegraphics[trim={0cm 0cm 0cm 0cm},clip,width=.48\textwidth]{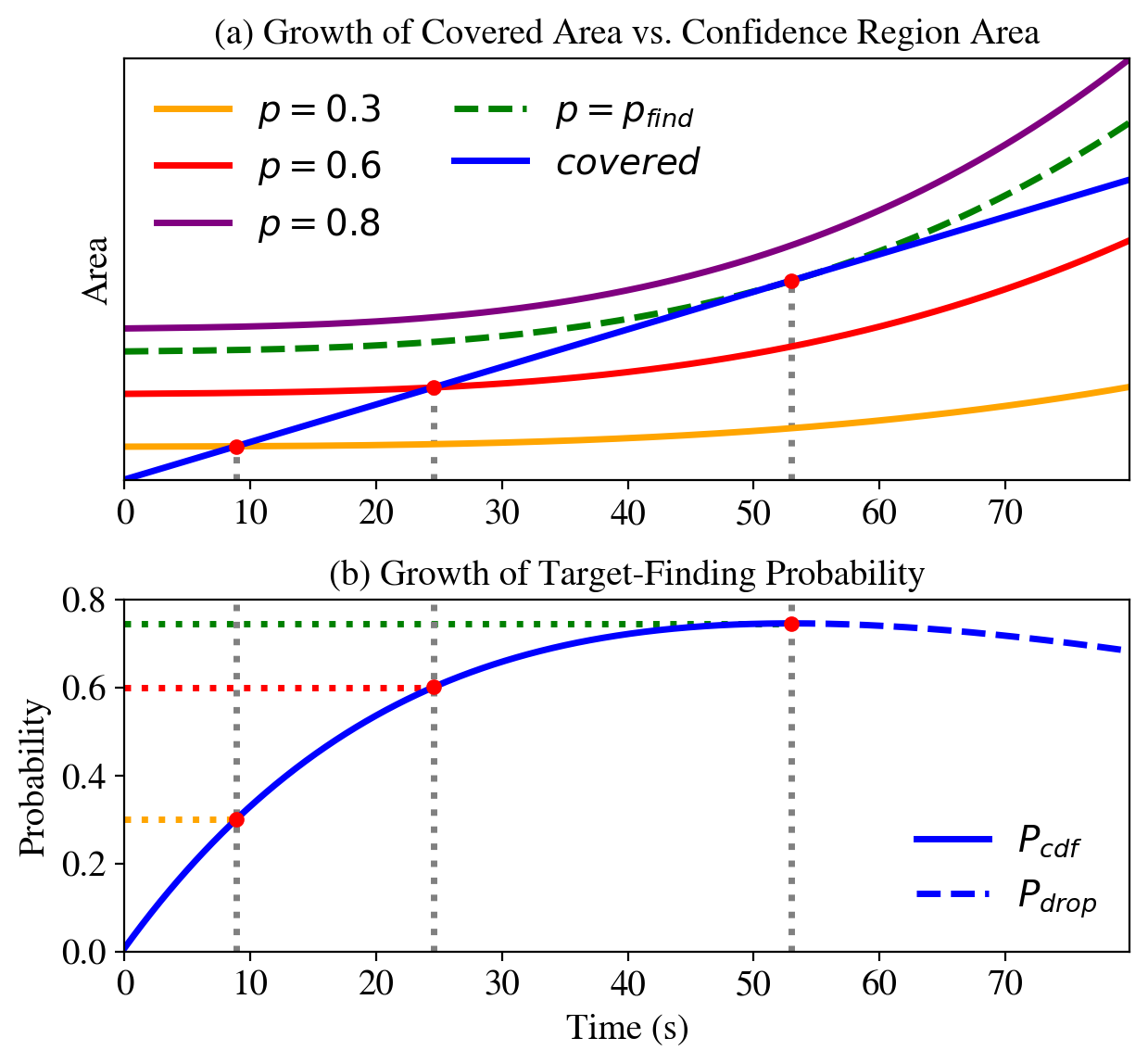} 
    \caption{An example of coverage result estimation. During the coverage, (a) shows that the covered area catches up the ellipse areas of some example confidence levels $p=\{0.3, 0.6, 0.8\}$, eventually becoming tangent to the highest attainable probability $p_\text{find}$. (b) illustrates the growth of target-finding probability. Both plots share the same time axis, and the intersection points in (a) correspond to the respective probabilities in (b). In this example, $t_\text{cutoff}=53.0$ and $p_\text{find}=0.746$ are determined by the reach of maximum target-finding probability. }
    \label{fig:prob_catch}
    \vspace{-0.3cm}
\end{figure}

The estimator is computed in a loop of increasing time step $k$. 
At each time step $k$, we approximate the covered area $S_\text{cover}$ as increasing linearly with time: 
\begin{align}
    S_\text{cover} = v_{a}w * k \tau + S_{\text{init}},
\end{align}
where $S_{\text{init}}$ is the sensor footprint area that lies at the center when the coverage begins.
Given the updated covariance of the target $\Sigma^{xy}(k)$ using Eq. \eqref{eq:prediction_model}, the probability of the target residing in the covered area, $P_\text{cdf}(k)$, is calculated using the chi-squared function:
\begin{align}
    &P_\text{cdf}(k) = \chi^2_{\text{cdf}}\left(\frac{S_\text{cover}}{\pi \sqrt{D}}, df=2 \right),\\
    &D = \text{det}(\Sigma^{xy}(k)).
\end{align}
$P_\text{cdf}(k)$ is an approximation of the actual probability of finding the target, as we assume that 
(1) the covered area closely resembles an elliptic region and 
(2) the probability of the target re-entering the area covered in the previous time step is low. 
The growth of $P_\text{cdf}(k)$ indicates an increasing probability of finding a target, as illustrated in Fig.~\ref{fig:prob_catch}. 

Meanwhile, as the target belief propagates according to \eqref{eq:prediction_model}, the state covariance continues to increase, resulting in a quadratic growth in uncertainty ellipses. Consequently, at a certain time step, the growth rate of an ellipse will surpass the linear coverage rate. At this point, an increase in coverage time $k$ will no longer contribute to an increase in $P_\text{cdf}(k)$, indicating that the maximum probability $p_\text{max}$ is reached, as illustrated in Fig.~\ref{fig:prob_catch}(b). Once this occurs, we assign the time $k$ to $t_\text{cutoff}$. 

The proposed coverage estimator enables the formulation of the task as an MDP planning problem, making it possible to develop a global planner based on the prediction of future coverage outcomes.

\subsection{Global Planner}
In addition to the coverage planner and its performance estimator, we developed a global planner that determines target visitation order in a dynamic environment, where the uncertainty regions evolve over time.
To minimize overall target uncertainty at budget depletion time, we developed a Monte Carlo tree search-based method in Markov decision process framework.

\subsubsection{Markov Decision Process}
We decompose the multi-target tracking problem into multiple sub-tasks. In each sub-task, if the agent chooses to pursue target $a$, the agent will intercept the state estimate of the target and start executing the coverage path described in \ref{sec: coverage path planner} until the target is detected, or $t_\text{cutoff}$ is reached. With possible coverage outcome of finding or missing the target, formulated as $o \in \{ \text{find}, \text{miss} \}$, the agent will decide the next target to pursue and execute the next sub-task accordingly.

We define the history $h_t$ as a sequence of actions and coverage outcome, $h_t = [a_1, o_1, a_2, o_2, \dots, a_t, o_t]$. From the available targets after $h_t$, the next target to pursue is selected as an action $a_{t+1}$, and the probability of finding this next target follows the output of the coverage estimator:
\begin{equation}
\begin{split}
&p_\text{find} = P(o_{t+1}=\text{Find} | h_t, a_{t+1}) = p_{\text{max}} \\
&p_\text{miss} = P(o_{t+1}=\text{Miss} | h_t, a_{t+1}) = 1 - p_{\text{max}}
\end{split}
\end{equation}

The action with its outcome will form a new coverage outcome and hence extend the history $h_{t+1}$ with the new states $a_{t+1}$ and $o_{t+1}$.
 
\subsubsection{Node Expansion}\label{sec: node expansion}
A state node $\mathbf{N}^s$ is characterized by a history $h$, the action list $\mathcal{A}$ consisting of all available targets, the remaining budget, agent pose, and the targets' belief.
With an action $a \in \mathcal{A}$, the state node can expand to generate an action node $\mathbf{N}^a$. 

The action node represents a sequence of actions where the agent intercepts a target's estimated position and carries out the spiral coverage search path with a fixed speed $v_a$. 
The outcome of the action is estimated using the coverage estimator, including the probability of detection and the expected time spent. 
When the remaining budget is lower than the expected time, the probability of detection is adjusted using the reduced coverage time.

\begin{algorithm}[t]
\caption{\textsc{MonteCarloPlanning}}
\label{alg: Simulation}
\SetInd{0.4em}{0.8em}
\DontPrintSemicolon
\small

\SetKwInOut{Input}{Input}
\SetKwInOut{Output}{Output}
\SetKwInOut{Config}{Config}

\SetKwFunction{Simulate}{Simulate}
\SetKwFunction{Rollout}{Rollout}
\SetKwFunction{TreeSearch}{TreeSearch}

\SetKwProg{Fn}{Procedure}{}{}

\Fn{$\textsc{TreeSearch}(\textnormal{IsReplan})$}
{
    \Repeat{$\textsc{Timeout}()$}
    {
        \If{$h = \text{empty}$}
        {
            $\mathbf{N}^s_\text{start} \leftarrow \textsc{Initialize}()$\;
        }
        \Else
        {
            $\mathbf{N}^s_\text{start} \leftarrow \textsc{CurrentState}()$\;
        }
        $\textsc{Simulate}(\mathbf{N}^s_\text{start})$\;
    }
    \If{$\text{IsReplan} = True$}
    {
        $\mathbf{N}^a \leftarrow \mathbf{N}^s_\text{start}.\textsc{PursuingTarget}()$\;
        $\mathbf{N}^s_\text{find} \leftarrow \textsc{GetStateNode}(\mathbf{N}^a, \text{Find})$\;
        $\mathbf{N}^s_\text{miss} \leftarrow \textsc{GetStateNode}(\mathbf{N}^a, \text{Miss})$\;
        \Return $\arg\max\limits_{b \in \mathbf{N}^s_\text{find}.\mathcal{A}} V(\mathbf{N}^a_\text{b}), 
                 \arg\max\limits_{b \in \mathbf{N}^s_\text{miss}.\mathcal{A}} V(\mathbf{N}^a_\text{b})$\;
    }
    \Return $\arg\max\limits_{b \in \mathbf{N}^s_\text{start}.\mathcal{A}} V(\mathbf{N}^s_\text{start})$\;
}

\vspace{0.5em} 


\vspace{0.5em} 

\Fn{$\textsc{Simulate}(\mathbf{N}^s)$}
{
    \If{$\mathbf{N}^s.\text{remain\_budget} \leq 0$}
    {
        \Return $U(\mathbf{N}^s)$\;
    }
    \If{$\mathbf{N}^s.\textsc{FirstTimeVisited}()$}
    {
        \For{$a \in \mathbf{N}^s.\mathcal{A}$}
        {
            $\mathbf{N}^a \leftarrow \textsc{ExpandState}(\mathbf{N}^s, a)$\;
            $\mathbf{N}^s.\textsc{AddChild}(\mathbf{N}^a)$
        }
        \Return $\textsc{Rollout}(\mathbf{N}^s)$\;
    }
    $a \leftarrow \arg \max \limits_{b \in \mathbf{N}^s.\mathcal{A}} V(\mathbf{N}^a_b) + c \sqrt{\frac{\log N(\mathbf{N}^s)}{N(\mathbf{N}^a_b)}}$\;
    $\mathbf{N}^a \leftarrow \textsc{GetActionNode}(\mathbf{N}^s, a)$\;
    $\text{outcome} \sim \text{Bernoulli}(\mathbf{N}^a.p_\text{find})$\;
    \If{$\mathbf{N}^a.\textsc{NotExpanded}(\text{outcome})$}
    {$\mathbf{N}^s_\text{next} \leftarrow \textsc{ExpandAction}(\mathbf{N}^a, \text{outcome})$}
    \Else
    {$\mathbf{N}^s_\text{next} \leftarrow \textsc{GetStateNode}(\mathbf{N}^a, \text{outcome})$}
    $U \leftarrow \textsc{Simulate}(\mathbf{N}^s_\text{next})$\;
    $N(\mathbf{N}^s) \gets N(\mathbf{N}^s) + 1$\;
    $N(\mathbf{N}^a) \leftarrow N(\mathbf{N}^a) + 1$\;
    $V(\mathbf{N}^a) \gets V(\mathbf{N}^a) + \frac{-U - V(\mathbf{N}^a)}{N(\mathbf{N}^a)}$\;
    \Return $U$\;
}
\end{algorithm}

The action node can expand to a new state node given an action's outcome according to $p_\text{find}$. 
Based on the expected coverage outcome and time spent, 
the new state node is updated with the remaining budget, the targets' beliefs, and the agent pose, which we set to the position estimate of the pursued target after the expected coverage time.
A target is removed from the action list $\mathcal{A}$ if it is not found.
In the case of consecutive identical actions, the agent follows the last observed target until budget depletion. In this case, this target's detection is guaranteed.

\subsubsection{Monte-Carlo Tree Search in MDP}
With the described node expansion process, we modified the Monte-Carlo planning method \cite{silver2010monte} to expand the tree during planning time. Such expansion can be either the first planning at the beginning of the task, or a replanning process while pursuing a target, as described in Algorithm~\ref{alg: Simulation}.

\begin{figure}[t]
    \centering
    \includegraphics[trim={0cm 0cm 0cm 0cm},clip,width=.48\textwidth]{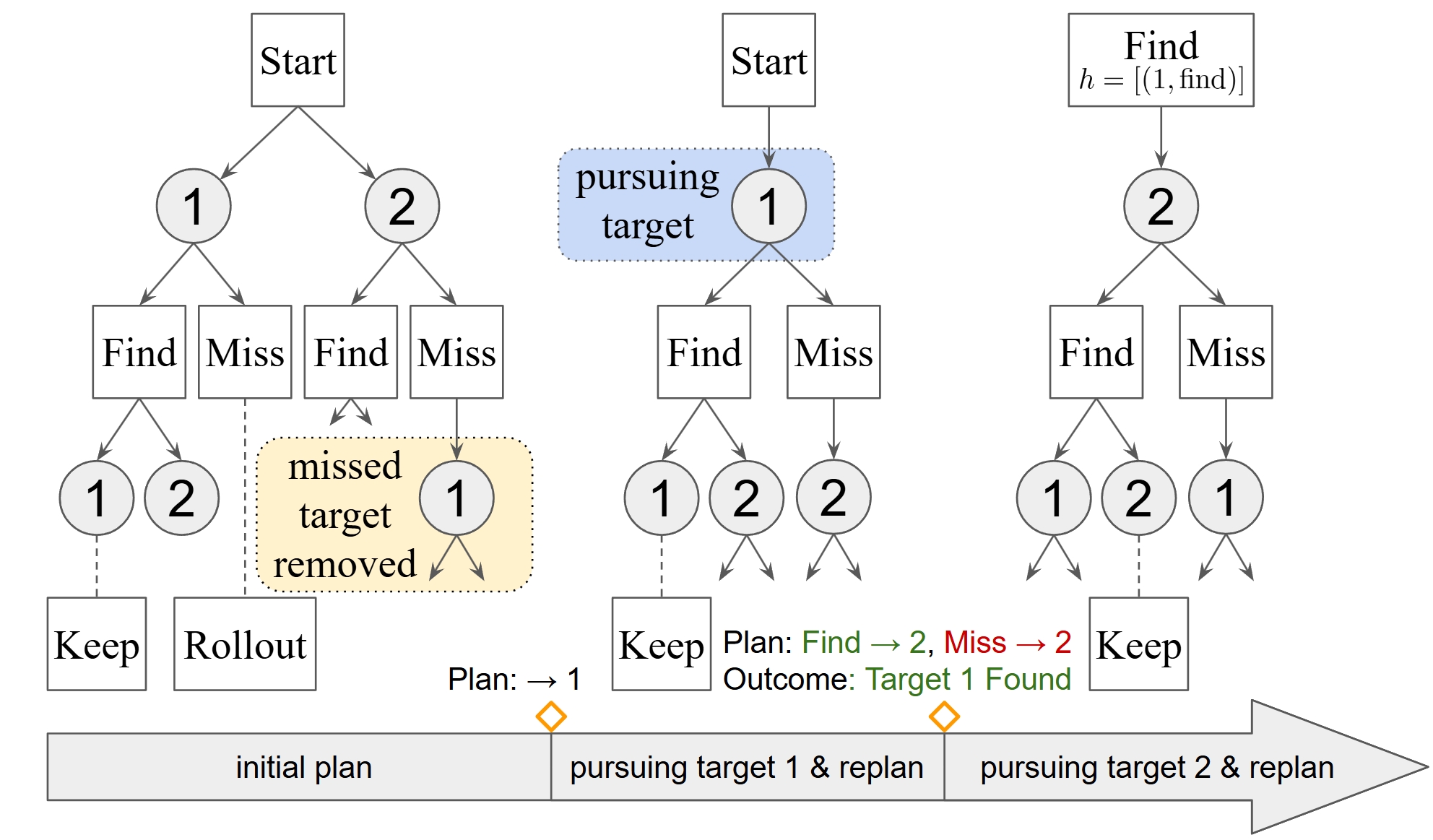} 
    \caption{Planning process example with 2 targets. The initial plan selects target 1 first. During the search for target 1, the system replans with the current action fixed, generates a conditional policy based on the coverage outcome, and repeats this replanning process for subsequent targets until the budget is depleted.}
    \label{fig: tree}
    \vspace{-0.2cm}
\end{figure}

The main simulation procedure is described in the function $\textsc{SIMULATE}$ in Algorithm \ref{alg: Simulation}. When the algorithm encounters a state node for the first time after adding it to the tree, it expands the node to initialize the next action nodes and performs a rollout with a random policy from the state to obtain a rollout value (line 18-22). Otherwise, an action is selected by using UCB1 \cite{auer2002finite} and an outcome is sampled according to $p_\text{find}$ in the corresponding action node (line 23-25), with which the next state node is either reached or initialized by extension (line 26-29). Such a simulation process is executed recursively, allowing action values to be updated through upward propagation, while the visitation count of the corresponding node is incremented by one after each simulation (line 30-33). The end of simulation along a branch is reached when the remaining budget is depleted, and the overall uncertainty $U$ of the depletion state is returned (line 16-17). Repetitive execution of the simulation from the start node enables the tree to keep expanding, with action values updated over time before it is terminated with timeout.

\subsubsection{Re-Planning}
The tree search is run repeatedly to determine the next target to pursue during replanning. After the initial plan is generated for the first target, replanning begins immediately, with the first action fixed to the pursued target. It will return a conditional plan based on the coverage outcome as described in line 11 and 12 of Algorithm~\ref{alg: Simulation} and Fig.~\ref{fig: tree}. The time limit for a replanning process is the actual coverage time, either after missing the target at $t_\text{cutoff}$ or finding the target before that, and it can also be capped by a limit as defined in Section~\ref{sec: experiment}.

\definecolor{pltRed}{RGB}{255, 5, 5}
\definecolor{pltYellow}{RGB}{200, 200, 0}
\definecolor{pltOrange}{RGB}{255, 171, 34}
\definecolor{pltBlue}{RGB}{26, 26, 255}
\definecolor{pltPurple}{RGB}{128, 0, 128}

\begin{figure}[t]
     \centering
     \begin{subfigure}[]{0.95\columnwidth}
         \centering
         \includegraphics[trim={0cm 0cm 0.2cm 0cm},clip,width=\textwidth]{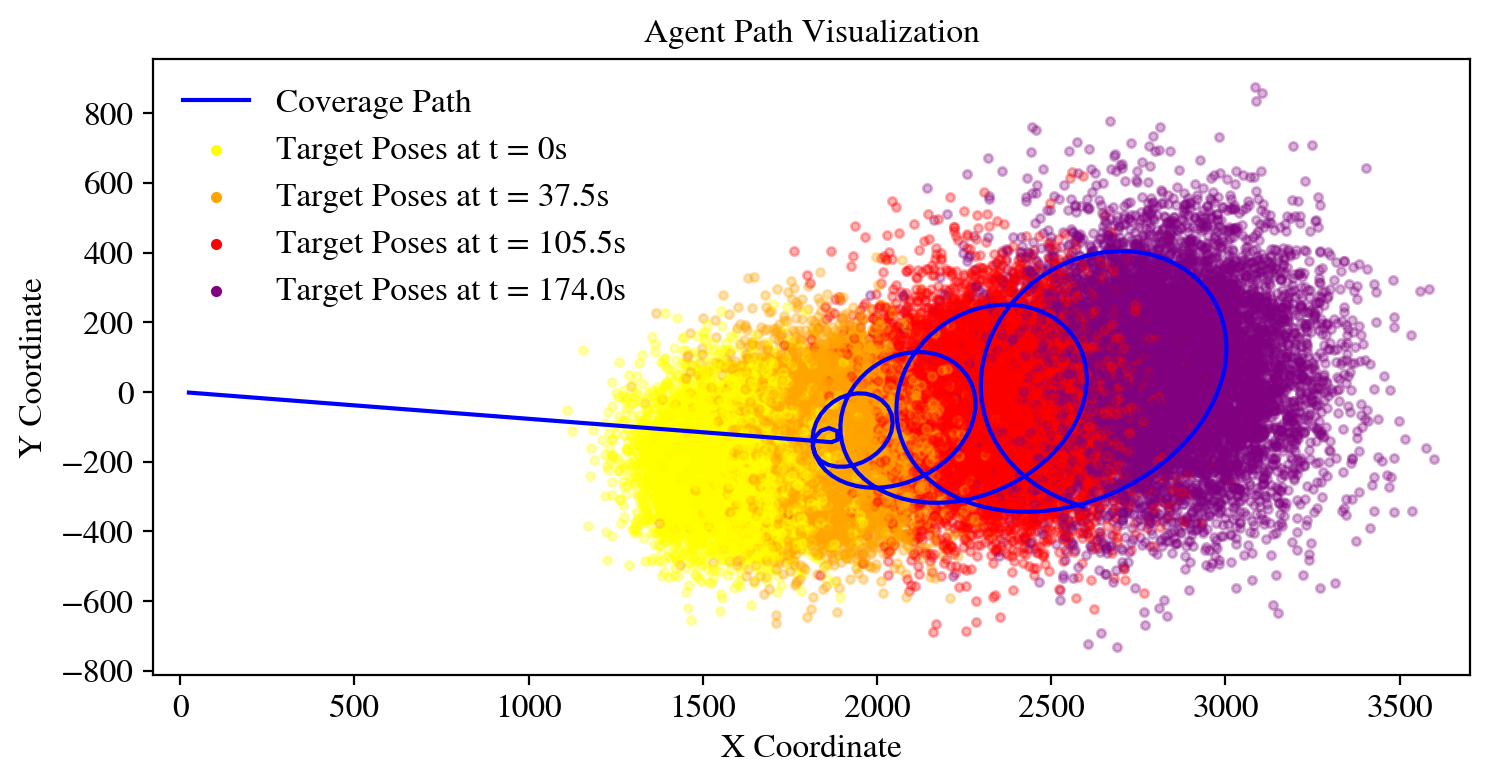}
         \caption{}
         \label{fig: coverage path}
     \end{subfigure}
    
     \hfill
     \begin{subfigure}[]{0.95\columnwidth}
         \centering
         \includegraphics[trim={0.2cm 0cm 0cm 0cm},clip,width=\textwidth]{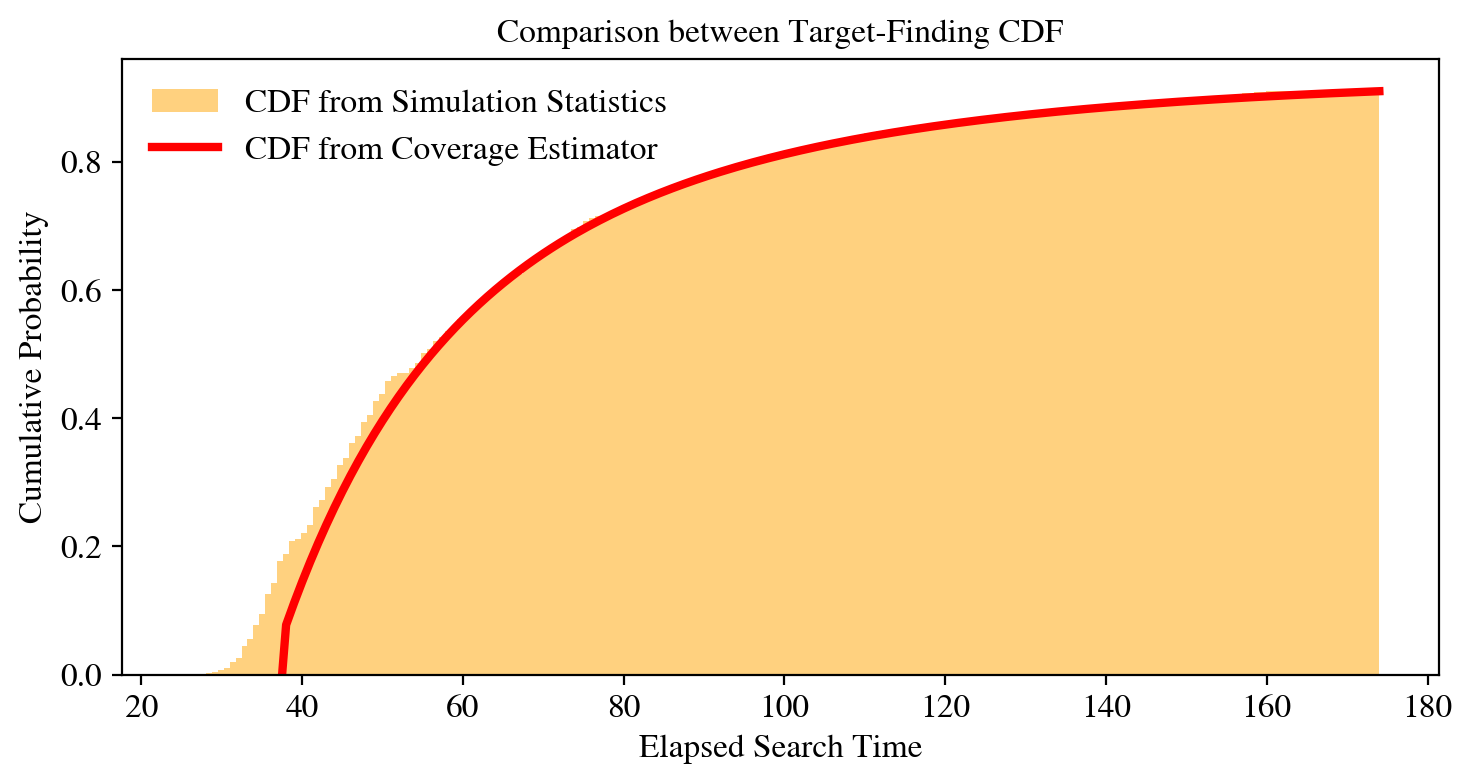}
         \caption{}
         \label{fig: estimation result}
     \end{subfigure}

    \caption{An example of a 10000-simulation setup and statistics. Fig. \ref{fig: coverage path}: the agent’s coverage path is fixed and target trajectories in each simulation are sampled and propagated with belief.
    Fig. \ref{fig: estimation result}: the comparison between the estimated target-finding CDF and simulation statistics.}
    \label{fig: coverage test}
    \vspace{-0.3cm}
\end{figure}

\section{Experiment} \label{sec: experiment}
To evaluate our proposed methods in comparison with related works, we conducted a series of experiments in simulated environments. We also carried out real-world experiments using a quadrotor.

In simulated testing environments, we first verified the accuracy of the proposed probability estimation algorithm for the spiral coverage planner, then we tested our planning pipeline with comparison to some existing methods for active target tracking tasks.
The time step of all experiments are set as $\tau = 0.5 \ \mathrm{s}$.
The experiments are run on an Intel 13900K processor, each simulation running on one logical P-core. 

\subsection{Simulated Experiments: Coverage Planner}

With coverage of a dynamic belief area being the crucial sub-task of the pipeline, it is important to estimate its outcome accurately to make proper decisions. We tested the coverage planner and validated the accuracy of the corresponding estimator through multiple simulations in single-target environments.

For each simulation, we initialize an agent with a start pose at $(0, 0)$, a speed of $30\ \mathrm{m/s}$ and the sensor width of $100\ \mathrm{m}$.
In each test case, the targets' initial state and motion noise are randomized as follows: 
the initial position $x_{i,0}$ and $y_{i,0}$ are sampled from a uniform distribution within $[-1000, 1000]\ \mathrm{m}$, while the initial velocity $v^x_{i,0}$ and $v^y_{i,0}$ are sampled uniformly from $[-3, 3]\ \mathrm{m/s}$. 
The initial uncertainty is described by the covariance matrix $\Sigma_{i, 0}$, where the position variances in the $x$ and $y$ directions are independently and uniformly sampled from $[500, 2000]$, and velocity variances are uniformly sampled from $[0.25, 0.5]$.
The correlation coefficient $\rho_{i, 0}$ between $x_{i, 0}$ and $y_{i, 0}$ is drawn uniformly from $[-0.8, 0.8]$ to set the off-diagonal elements for pose in $\Sigma_{i, 0}$. We also set
\begin{align*}
 F = 
\begin{bmatrix}
I_2 & \tau I_2 \\
0 & I_2 
\end{bmatrix}, ~
 Q_i = q
\begin{bmatrix}
\tau^3/3 I_2 & \tau^2/2 I_2\\
\tau^2/2 I_2 & I_2
\end{bmatrix}
\end{align*}
with q uniformly sampled from $[0.0001, 0.0005]$.

In actual multi-target tracking tasks, an agent may delay the pursuit of a specific target while prioritizing others.
To model this, we set a pseudo departure time $t_\text{d} \in \{ 0, 100, 200 \}$ seconds, during which the agent remains stationary, while the target moves and the belief propagates. After such delay, the agent plans and executes a path that first intercepts the belief center and then covers it with the shifting spiral path as described in \ref{sec: coverage path planner}.

For each test case and $t_\text{d}$, we conducted 10000 runs, and in each run, the target path is different with the initial pose and motion noises drawn from the distribution, while the agent's path from \ref{sec: coverage path planner} and the estimated coverage outcome from \ref{sec: coverage estimation} remains identical.
An example test case is shown in Fig.~\ref{fig: estimation result}, where the estimation curve of $P_\text{cdf}$ closely matches the actual probability distribution of finding the target.
At the initial stage of search, the estimator underestimates the actual probability of finding a target, due to the fact that some targets are found before the spiral coverage begins as the agent travels on its way to the most probable target location.

We tested through 1000 randomly generated single-target cases, 
and compared the empirical statistics from simulations with the results from the coverage estimator.
We present our results in Table~\ref{table: coverage result}. Both the estimated probability of finding the target and the expected target-finding time closely match the empirical statistics obtained from simulations, as demonstrated by the low mean absolute percentage error (MAPE) values. 
Hence, the estimated probability and time can be used as a reliable estimate to generate new state nodes in the global planner.

\begin{table}[t]
    \centering
    \vspace{3mm}
    \resizebox{1\columnwidth}{!}
    {
    \renewcommand{\arraystretch}{1.5}
    \begin{tabular}{|c||*{3}{c|}}
        \hline \xrowht{10pt}
        \makebox{}
        &\makebox[5em]{$t_\text{d}=0$s}&\makebox[5em]{$t_\text{d}=100$s}&\makebox[5em]{$t_\text{d}=200$s}\\
        \hhline{|=||=|=|=|} \xrowht{10pt}
        $\text{MAPE}(P_\text{find})$ & $3.58\%$ & $2.46\%$ & $5.46\%$\\
        \hline \xrowht{10pt}
        $\text{MAPE}(\mathbb{E}[T_\text{find}])$ & $5.18\%$ & $3.42\%$ & $3.73\%$\\
        \hline
    \end{tabular}
    }
    \caption{~ Comparison of estimated and empirical probabilities of finding the target $P_\text{find}$ and expected target-finding time $\mathbb{E}([T_\text{find}])$ using Mean Absolute Percentage Error (MAPE).}
    \label{table: coverage result}
    \vspace{-0.3cm}
\end{table}

\subsection{Simulated Experiments: Multi-Target Tracking}

We validated our integrated target-tracking planner using randomized multi-target tracking tasks. 
For each $n \in \{2, 3, 4, 5, 6\}$, we randomize 100 cases with $n$ targets.
Each planner was evaluated once per scenario, and we recorded the final uncertainty $U$ upon budget depletion. 
We set the total budget $B = 900$ seconds, and set the planning time limit 30 seconds for the tree search.

We compared our method against ARVI \cite{8260881}, A* with information heuristics \cite{8968147}, and an ADFQ-based learning approach \cite{10.5555/3367243.3367402}.
The results are presented separately due to differing assumptions: ARVI and A* require identical target models and initial covariances, while ADFQ supports a heterogeneous target setup.

\subsubsection{Comparison with ARVI and A* with Heuristic}
\begin{figure}[t]
    \centering
    \includegraphics[trim={0cm 0cm 0cm 0cm},clip,width=\columnwidth]{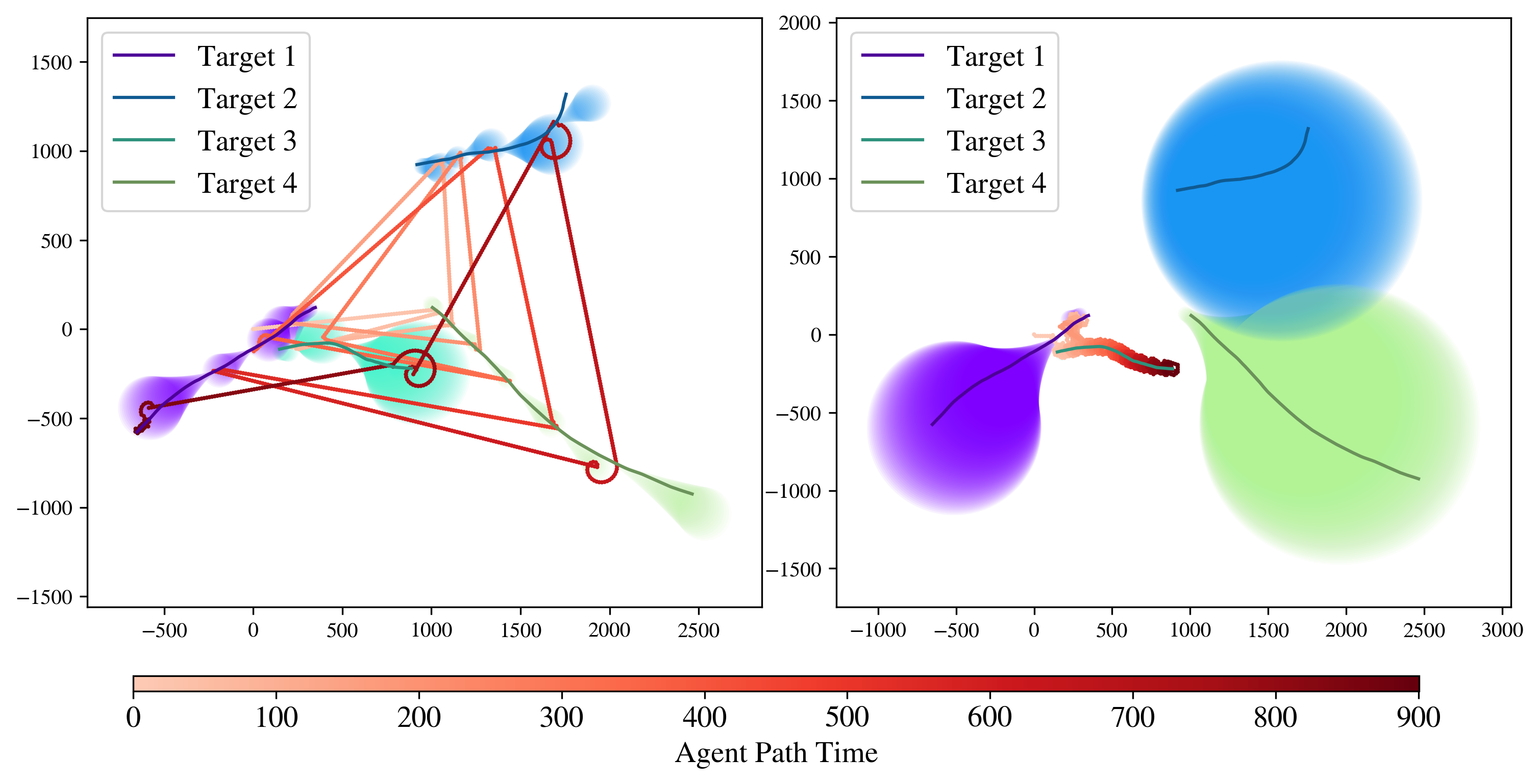} 
    \caption{Comparison between our method (left) and the ARVI planner (right) on a representative test case. Each subplot shows the agent trajectory (colored by time) and target trajectories with evolving belief ellipses. When targets are not observed, their belief regions expand, and the beliefs shrink accordingly upon detection. Our method continuously explores and updates all targets' beliefs, maintaining global awareness. In contrast, ARVI focuses solely on targets 3 and 4, ultimately fixating on target 3, allowing the uncertainties of the remaining targets to grow unbounded.}
    \label{fig: qualitative compare}
    \vspace{-0.3cm}
\end{figure}

To match the optimization object with our formulation, we modified the cost function of ARVI \cite{8260881} and A*\cite{8968147}, together with the heuristic of A* \cite{8968147} from the sum of overall uncertainty in each step to the difference in uncertainty between steps. Qualitatively, we verified this change with small environments, where the ARVI agent showed reasonable behavior of moving back and forth between the centers of belief areas to minimize the overall uncertainty. However, with a larger environment and sparser target belief areas, the agent usually reaches the first few targets and maintains track of one, leaving the uncertainty of all other targets growing for the entire mission duration, 
as illustrated in Fig. \ref{fig: qualitative compare}.

The quantitative simulation results are presented in Table~\ref{table: global compare 1}. These results demonstrate that our planner almost consistently outperforms these existing planners in reducing the overall uncertainty of environments upon budget depletion. Depending on the number of targets, our method achieves a reduction in uncertainty ranging from $12.22\%$ to $65.58\%$ compared to the ARVI planner and a reduction ranging from $19.60\%$ to $69.82\%$ from the A* with heuristic across all tested scenarios. The baseline methods struggle with large-scale tasks, as they lack the ability to reason about when and how to detect targets whose belief regions have expanded beyond the sensor's field of view. In contrast, while our proposed planner outperforms these baselines, its advantage diminishes as the number of targets increases. This is due to the growing complexity of the problem, which limits the number of visits to each node in the MCTS tree and may lead to suboptimal planning decisions under the planning time constraint.
\subsubsection{Comparison with Learning Methods}
To compare against the ADFQ based learning method for target tracking, we implement the test environment similar to \cite{jeong2021deep}, and modify the agent motion model to allow holonomic motion. 

The quantitative simulation results are presented in Table~\ref{table: global compare 2}. These results demonstrate that the learning based methods struggle in large environments with sparse targets and long-horizon planning. We hypothesize that the large distances between targets in such tasks introduce long-term dependencies and sparse rewards, which can pose significant challenges for learning-based models, especially when the task is not decomposed into sub-tasks that allow the agent to focus on a single target at a time. Depending on the number of targets, our method achieves a reduction in uncertainty ranging from $11.40\%$ to $29.53\%$  compared to the ADFQ-based planner across all tested scenarios.

\subsection{Real-World Experiment}
To verify the deployability of our method, we also run our planning system on a real UAV platform and validated its performance in an open environment with two pseudo targets, as illustrated in Fig. \ref{fig: full_overview}. We use pseudo targets with an initial separation of $20$m and moving at a velocity of $3$m/s. To simplify the setup and focus on verifying the planning algorithm, we assume target detection as soon as the targets lie within the camera's field of view. In this experiment, the drone demonstrates the behavior that keeps alternating between two targets, updating their belief, thereby maintaining the overall uncertainty at a reasonable level.
The experiment verifies that the proposed approach generates realistic robot motion and has potential for more complex real-world missions with onboard sensors.


     


\begin{table}
    \centering
    \vspace{3mm}
    \resizebox{1\columnwidth}{!}
    {
    \renewcommand{\arraystretch}{1.5}
    \begin{tabular}{|c||*{5}{c|}}
        \hline \xrowht{15pt}
        \makebox{Planner}
        &\makebox[4.5em]{$n=2$}&\makebox[4.5em]{$n=3$}&\makebox[4.5em]{$n=4$}&\makebox[4.5em]{$n=5$} &\makebox[4.5em]{$n=6$} \\
        \hhline{|=||=|=|=|=|=|} \xrowht{15pt}
        ARVI & ${40.64} \pm 11.58$ & ${62.84} \pm 12.69$ & \makecell[c]{${83.55} \pm 13.30$} & \makecell[c]{${105.51} \pm 15.22$} & \makecell[c]{${127.71} \pm 15.25$}\\
        \hline \xrowht{15pt}
        Heuristic & \makecell[c]{${46.35} \pm 1.52$} & \makecell[c]{${69.42} \pm 2.24$} & ${93.05} \pm 2.81$ & \makecell[c]{${115.84} \pm 3.65$} & \makecell[c]{${139.42} \pm 3.57$}\\
        \hline \xrowht{15pt}
        \bf Ours & \makecell[c]{$\bm{13.99} \pm 3.97$} & $\bm{26.25} \pm 7.03$ & \makecell[c]{$\bm{51.47} \pm 17.07$} & \makecell[c]{$\bm{83.59} \pm 21.33$} & \makecell[c]{$\bm{112.10} \pm 17.62$}\\
        \hline
    \end{tabular}
    }
    \caption{~ Comparison of mean $\pm$ standard deviation overall uncertainty $U$ over 1000 randomized cases for different target number $n$ after $900$ seconds tracking with planner algorithms including ARVI, A* with heuristic and ours.}
    \label{table: global compare 1}
    \vspace{-3.5mm}
\end{table}

\begin{table}
    \centering
    \vspace{3mm}
    \resizebox{1\columnwidth}{!}
    {
    \renewcommand{\arraystretch}{1.5}
    \begin{tabular}{|c||*{5}{c|}}
        \hline \xrowht{15pt}
        \makebox{Planner}
        &\makebox[4.5em]{$n=2$}&\makebox[4.5em]{$n=3$}&\makebox[4.5em]{$n=4$}&\makebox[4.5em]{$n=5$} &\makebox[4.5em]{$n=6$} \\
        \hhline{|=||=|=|=|=|=|} \xrowht{15pt}
        ADFQ & ${50.90} \pm 1.47$ & ${75.76} \pm 2.72$ & \makecell[c]{${101.14} \pm 3.02$} & \makecell[c]{${126.22} \pm 2.88$} & \makecell[c]{${152.10} \pm 3.05$}\\
        \hline \xrowht{15pt}
        \bf Ours & \makecell[c]{$\bm{35.87} \pm 6.87$} & \makecell[c]{$\bm{59.77} \pm 7.15$} & $\bm{83.38} \pm 7.93$ & \makecell[c]{$\bm{108.61} \pm 8.06$} & \makecell[c]{$\bm{134.76} \pm 7.76$}\\
        \hline
    \end{tabular}
    }
    \caption{~ Comparison of mean $\pm$ standard deviation overall uncertainty $U$ over 1000 randomized cases for different target number $n$ after $900$ seconds tracking with planner algorithms including ADFQ and ours.}
    \label{table: global compare 2}
    \vspace{-3.5mm}
\end{table}



\section{Conclusion} \label{sec:conclusion}


We proposed a hierarchical planning pipeline for long-horizon multi-target tracking over a large workspace. Our approach decomposes the multi-target tracking task into a sequence of sub-tasks. We made such decomposition possible by addressing key challenges, including the coverage planning for an evolving belief area for each sub-task and the estimation of coverage outcome, enabling the multi-agent tracking task to be re-formatted as an MDP problem that we solved with an MCTS-based global planner. Our experimental results validate the effectiveness of the proposed approach in improving target reacquisition and reducing track uncertainty. 

Our future work will focus on developing coverage planners for a broader range of motion models and extending the outcome estimator to achieve predictability across diverse target dynamics. In addition, we plan to investigate scalability improvements to handle larger target sets and explore extensions that explicitly consider occlusions and environmental constraints, such as structured or cluttered spaces, to further enhance real-world applicability.






\bibliography{ref}

\end{document}